\documentclass[runningheads]{llncs}
\usepackage[T1]{fontenc}
\usepackage{graphicx}
\usepackage{graphicx}
\usepackage{booktabs}
\usepackage{color}
\usepackage{booktabs}
\usepackage{makecell}
\usepackage{bbding}

\usepackage{ulem}

\usepackage{array}

\usepackage{overpic}
\usepackage{bm}
\usepackage{amsmath,array,arydshln}
\usepackage{multicol}
\usepackage{multirow}
\usepackage{hyperref}
\usepackage{amsfonts}
\usepackage{cite}
\usepackage{ifsym}
\usepackage{subfigure}
\usepackage{hyperref}
\newcommand{\ve}[1]{\mathbf{#1}}

\begin{document}
\title{Adapter Learning in Pretrained Feature Extractor for Continual Learning of Diseases}
%
%


\author{Wentao Zhang\inst{1,4}\thanks{Authors contributed equally.} \and 
Yujun Huang\inst{1,4}$^\star$ \and 
Tong Zhang\inst{2} \and 
Qingsong Zou\inst{1,3} \and 
Wei-Shi Zheng\inst{1,4} \and 
Ruixuan Wang\inst{1,2,4}\thanks{Corresponding author: wangruix5@mail.sysu.edu.cn}} 

%

\institute{School of Computer Science and Engineering, Sun Yat-sen Univerisity, Guangzhou, China  \and Peng Cheng Laboratory, Shenzhen, China \and Guangdong Province Key Laboratory of Computational Science, Sun Yat-sen University, Guangzhou, China \and Key Laboratory of Machine Intelligence and Advanced Computing, MOE, Guangzhou, China}

%

\maketitle              
\begin{abstract}
Currently intelligent diagnosis systems lack the ability of continually learning to diagnose new diseases once deployed, under the condition of preserving old disease knowledge. In particular, updating an intelligent diagnosis system with training data of new diseases would cause catastrophic forgetting of old disease knowledge. To address 
the catastrophic forgetting issue, {an \textbf{A}dapter-based \textbf{C}ontinual \textbf{L}earning framework called ACL}  
is proposed to help effectively learn a set of new diseases at each round (or task) of continual learning, without changing the shared feature extractor. 
The learnable lightweight task-specific adapter(s) can be flexibly designed (e.g., two convolutional layers) and then added to the pretrained and fixed feature extractor. Together with a specially designed task-specific head which absorbs all previously learned old diseases as a single `out-of-distribution' category, task-specific adapter(s) can help the pretrained feature extractor more effectively extract discriminative features between diseases. In addition, {a simple yet effective fine-tuning is applied to collaboratively fine-tune multiple task-specific heads such that outputs from different heads are comparable and consequently the appropriate classifier head can be more accurately selected during model inference}. Extensive empirical evaluations on three 
image datasets demonstrate the superior performance of {ACL} 
in continual learning of new diseases.  {The source code is available at \url{https://github.com/GiantJun/CL_Pytorch}}.
\end{abstract}

\section{Introduction}
Deep neural networks have shown expert-level performance in various disease diagnoses~\cite{zhou2021ensembled,zheng2022deep}. In practice, a deep neural network is often limited to the diagnosis of only a few diseases, partly because it is challenging to collect enough training data of all diseases even for a specific body tissue or organ. One possible solution is to enable a deployed intelligent diagnosis system to continually learn new diseases with collected new training data later. However, if old data are not accessible due to certain reasons (e.g., challenge in data sharing), current intelligent systems will suffer from catastrophic forgetting of old knowledge when learning new diseases~\cite{li2017learning}.

Multiple approaches have been proposed to alleviate the catastrophic forgetting issue. One approach aims to determine part of the model parameters which are crucial to old knowledge and tries to keep these parameters unchanged during learning new knowledge~\cite{kirkpatrick2017overcoming,lopez2017gradient,chaudhry2018efficient}. Another approach aims to preserve old knowledge by making the updated model imitate the behaviour (e.g., output at certain layer) of the old model particularly with the help of knowledge distillation technique~\cite{li2017learning,douillard2020podnet,zhao2020maintaining}. Storing a small amount of old data or synthesizing old data relevant to old knowledge and using them together with training data of new knowledge can often help significantly alleviate forgetting of old knowledge~\cite{buzzega2020dark,rebuffi2017icarl,boschini2022transfer,shin2017continual}. 
Although the above approaches can help the updated model keep old knowledge to some extent, they often fall into the dilemma of model plasticity (for new knowledge learning) and stability (for old knowledge preservation). 
In order to resolve this dilemma, new model components (e.g., neurons or layers in neural networks) can be added specifically for {learning new knowledge}, while old parameters are largely kept unchanged for old knowledge~\cite{verma2021efficient,yan2021dynamically,rusu2016progressive}. While this approach has shown state-of-the-art continual learning performance, it faces the problem of rapid model expansion and effective fusion of new model components into the existing ones. To alleviate the model expansion issue and meanwhile well preserve old knowledge, researchers have started to explore the usage of a pretrained and fixed feature extractor for the whole process of continual learning~\cite{wang2022learning,wang2022dualprompt,kim2022multi,yang2021continual}, where the challenge is to discriminate between different classes of knowledge with limited learnable parameters.

In this study, inspired by recent advances in transfer learning in natural language processing~\cite{houlsby2019parameter,ding2022delta}, we propose adding a light-weight learnable module called adapter to a pretrained and fixed convolutional neural network (CNN) for effective continual learning of new knowledge. For each round of continual learning, the CNN model will be updated to learn a set of new classes (hereinafter also called learning a new task). The learnable task-specific adapters are added between consecutive convolutional stages to help the pretrained CNN feature extractor more effectively extract discriminative features of new diseases. To the best of our knowledge, it is the first time to apply the idea of CNN adapter in the continual learning field.
In addition, to keep extracted features discriminative between different tasks, a special task-specific classifier head is added when learning each new task, in which all previously learned old classes are considered as the `out-of-distribution' (OOD) class and correspond to an additional output neuron in each task-specific classifier head. A simple yet effective fine-tuning strategy is applied to calibrate outputs between multiple task-specific heads.  Extensive empirical evaluations on three 
image datasets show that the proposed method outperforms existing continual learning methods by a large margin, consistently supporting the effectiveness of the proposed method.



\begin{figure}[t]
\includegraphics[width=1\linewidth]{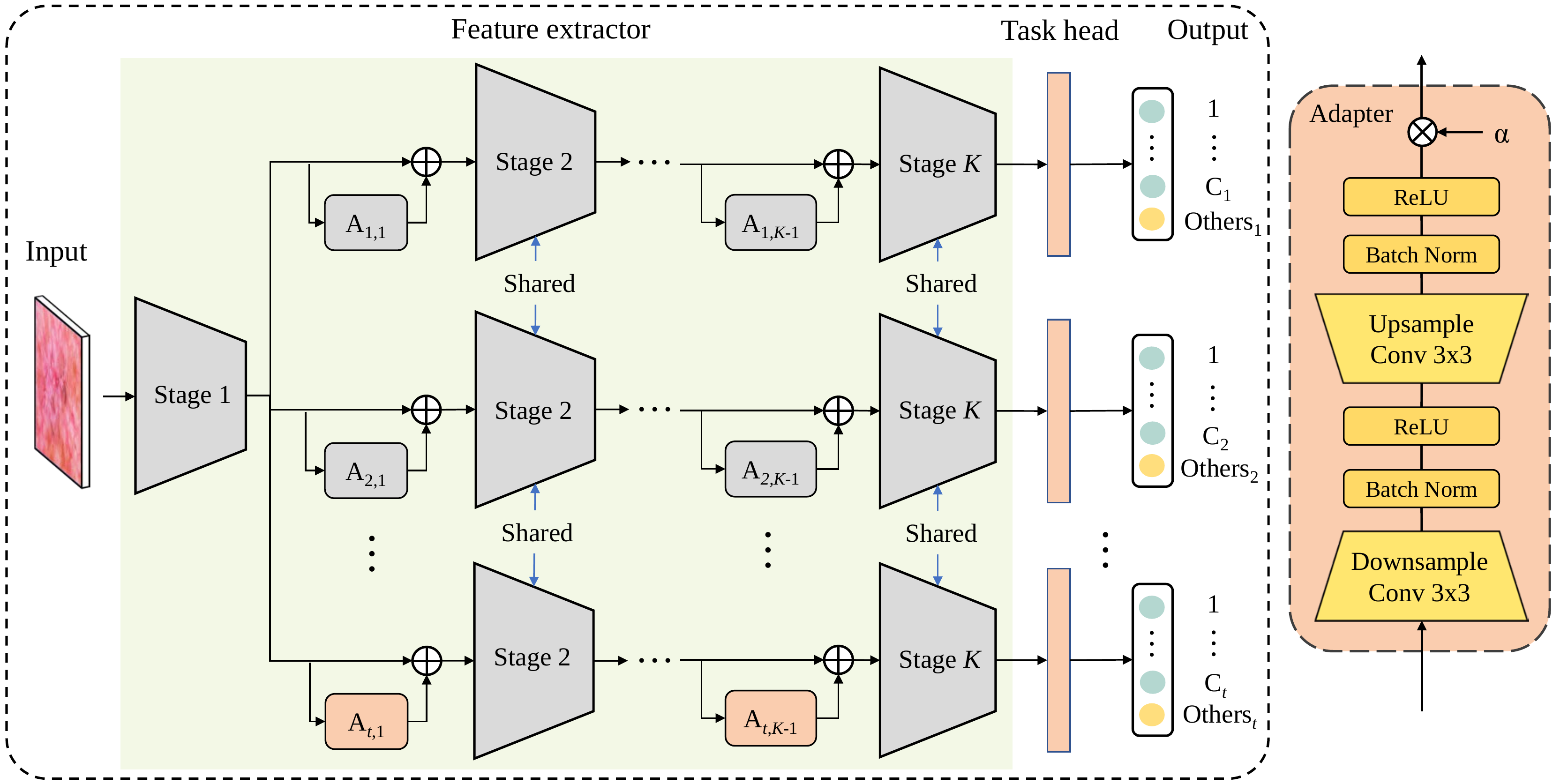}
\caption{{The proposed framework for continual learning of new diseases. Left: task-specific adapters ($\{A_{t,1}, A_{t,1}, \ldots, A_{t,K-1}\}$ in orange) between consecutive convolutional stages are added and learned for a new task. After learning the new task-specific adapters, all tasks' classifier heads (orange rectangles) are fine-tuned with balanced training data. 
The pretrained feature extractor is fixed during the continual learning process. 
Right: the structure of each adapter, with $\alpha$ representing the global scaling.}
} 
\label{fig:framework}
\end{figure}

\section{Method}

This study aims to improve continual learning performance of an intelligent diagnosis system. At each learning round, following previous studies~\cite{buzzega2020dark,boschini2022transfer} which show that rehearsal with old samples can significantly improve continual learning performance, the system will be updated based on the training data of new diseases and {preserved}  small subset for each previously learned disease. During inference, the system is expected to accurately diagnose all learned diseases, without knowing which round (i.e., task) the class of any test input is from. 

\subsection{Overall framework} 
We propose an {\textbf{A}dapter-based \textbf{C}ontinual \textbf{L}earning framework called ACL} with a multi-head training strategy. With the motivation to make full use of {readily available} pretrained CNN models and slow down the speed of model expansion that appears 
in some state-of-the-art continual learning methods (e.g., DER~\cite{yan2021dynamically}), and inspired by a recently developed transfer learning strategy Delta tuning for downstream tasks in natural language process~\cite{ding2022delta}, we propose adding a learnable light-weight adapter between consecutive convolutional stages in a pretrained and fixed CNN model when learning new classes of diseases at each learning round (Figure~\ref{fig:framework}). Each round of continual learning as a unique task is associated with task-specific adapters and a task-specific classifier head. Model update at each round of continual learning is to find optimal parameters in the newly added task-specific adapters and classifier head. During inference, since multiple classifier heads exist, the correct head containing the class of a given input is expected to be selected. 
{In order to establish the potential connections between tasks and further boost the continual learning performance}, a two-stage multi-head learning strategy was proposed by including the idea of out-of-distribution (OOD) detection.




\subsection{Task-specific adapters}
State-of-the-art continual learning methods try to preserve old knowledge by either combining old fixed feature extractors with the newly learned feature extractor~\cite{yan2021dynamically}, or by using a shared and fixed pretrained feature extractor~\cite{yang2021continual}. However, simply combining feature extractors over rounds of continual learning would rapidly expand the model, while using a shared and fixed feature extractor could largely limit model ability of learning new knowledge because only the model head can be tuned to discriminate between classes. To resolve this dilemma, we propose adding a light-weight task-specific module called adapter into a single pretrained and fixed CNN feature extractor (e.g., with the ResNet backbone), such that the model expands very slowly and the adapter-tuned feature extractor for each old task is fixed when learning a new task. In this way, old knowledge largely stored in the feature extractor and associated task-specific adapters will be well preserved when the model learns new knowledge at subsequent rounds of continual learning. 
Formally, suppose the pretrained CNN feature extractor contains $K$ stages of convolutional layers (e.g., 5 stages in ResNet), and the output feature maps from the $k$-th stage is denoted by $\ve{z}_k, k \in \{1,\ldots, K-1\}$. Then, when the model learns a new task at the $t$-th round of continual learning, a task-specific adapter $A_{t,k}$ is added between the $k$-th and $(k+1)$-th stages as follows,
\begin{equation}
\hat{\ve{z}}_{t,k} =  A_{t,k}(\ve{z}_k) + \ve{z}_k \,, 
\end{equation}
where the adapter-tuned output $\hat{\ve{z}}_{t,k}$ will be used as input to the $(k+1)$-th stage. The light-weight adapter can be flexibly designed. In this study, a simple two-layer convolution module followed by a global scaling is designed as the adapter (Figure~\ref{fig:framework}, Right). 
The input feature maps to the adapter are spatially downsampled by the first convolutional layer and then upsampled by the second layer.  
The global scaling factor $\alpha$ is learned together with the two convolutional layers. 
The proposed task-specific adapter for continual learning is inspired by Delta tuning~\cite{ding2022delta} which adds learnable 2-layer perceptron(s) into a pretrained and fixed {Transformer} model in natural language processing. 
Different from Delta tuning which is used as a transfer learning strategy to adapt a pretrained model for any individual downstream task, the proposed task-specific adapter is used as a continual learning strategy to help a model continually learn new knowledge over multiple rounds (i.e., multiple tasks) in image processing, with each round corresponding to a specific set of adapters. 

{Also note that the proposed adapter differs from existing adapters in CLIP-Adapter (CA)~\cite{gao2021clip} and Tip-Adapter (TA)~\cite{zhang2022tip}. First, in structure, CA and TA use 2-layer MLP or cache model, while ours uses a 2-layer convnet with a global scaling factor. Second, the number and locations of adapters in model are different. CA and TA use adapter only at output of the last layer, while ours appears between each two consecutive CNN stages. Third, the roles of adapters are different. Existing adapters are for few-shot classification, while ours is for continual learning. It is also different from current prompt tuning. Prompts appear as part of input to the first or/and intermediate layer(s) of model, often in the form of learnable tokens for Transformer or image regions for CNNs. In contrast, our adapter appears as an embedded neural module for each two consecutive CNN stages, in the form of sub-network.}


\subsection{Task-specific head}\label{section:task_head}
Task-specific head is proposed to alleviate the potential feature fusion issue in current state-of-the-art methods~\cite{yan2021dynamically} which combine task-specific features by a unified classifier head. In particular, if feature outputs from multiple task-specific feature extractors are simply fused by concatenating or averaging followed by a unified 1- or 2-layer percepton (as in~\cite{yan2021dynamically}), discriminative feature information appearing only in those classes of a specific task could become less salient after fusion with multiple (possibly less discriminative) features from other task-specific feature extractors. As a result, current state-of-the-art methods often require storing relatively more old data to help train a discriminative unified classifier head between different classes. To avoid the possible reduction in feature discriminability, we propose not fusing features from multiple feature extractors, but a task-specific classifier head for each task. Each task-specific head consists of one fully connected layer followed by a softmax operator. Specially, for a task containing $C$ new classes, one additional class absorbing all previously learned old classes (`others' output neuron in Figure~\ref{fig:framework}) is also included, and therefore the number of output elements from the softmax will be $C+1$. The `others' output is used to predict the probability of the input image being from certain class of any other task rather than from the current task. In other words, each task-specific head has the ability of out-of-distribution (OOD) ability with the help of the `others' output neuron. At the $t$-round of continual learning (i.e., for the $t$-th task learning), the task-specific adapters and the task-specific classifier head can be directly optimized, e.g., by cross-entropy loss, with the $C$ new classes of training data and the `others' class of all preserved old data.

However, training the task-specific classifier head without considering its relationship with existing classifier heads of previously learned tasks may cause the head selection issue during model inference. For example, a previously learned old classifier head may consider an input of latterly learned class as one of the old classes (correspondingly the `others' output from the old head will be low). In other words, the `others' outputs from multiple classifier heads cannot not be reliably compared (i.e., not calibrated) with one another if each classifier head is trained individually. In this case, if all classifier heads consider a test input as `others' class with high confidence or multiple classifier heads consider a test input as one of their classes, it would become difficult to choose an appropriate classifier head for final prediction. To resolve the head selection issue, after initial training of the current task's adapters and classifier head, all the tasks' heads are fine-tuned together such that all `others' outputs from the multiple heads are comparable.
In short, at the $t$-th round of continual learning, the $t$ task-specific classifier heads can be fine-tuned by minimizing the loss function $L$,
\begin{equation}
    L = \frac{1}{t} \sum_{s=1}^t L^c_{s}  \,, 
\end{equation}
where $L^c_{s}$ is the cross-entropy loss for the $s$-th classifier head. 
Following the fine-tuning step in previous continual learning studies~\cite{yan2021dynamically,castro2018end}, training data of the current $t$-th task are sub-sampled such that training data in the fine-tuning step are balanced across all learned classes so far. Note that for each input image, multiple runs of feature extraction are performed, with each run adding adapters of a different task to the original feature extractor and extracting the feature vector for the corresponding task-specific head. Also note that in the fine-tuning step, adapters of all tasks are fixed and not tuned. Compared to training the adapters of each task with all training data of the corresponding task, fine-tuning these adapters would likely cause over-fitting of the adapters to the  sub-sampled data and therefore is avoided in the fine-tuning step.

Once the multi-head classifier is fine-tuned at the $t$-th round of continual learning, the classifier can be applied to predict any test data as one of all the learned classes so far. 
First, the task head with the smallest `others' {output probability (among all $t$ ‘others’ outputs)} is selected, and then the class with the highest output from the selected task head is selected as the final prediction result. Although unlikely selected, the `others' class in the selected task head is excluded for the final prediction.

\section{Experiment results}



\subsection{Experimental setup}

{Four datasets were used to evaluate the proposed {ACL}
(Table~\ref{tab1}).} Among them,
Skin8 is imbalanced across classes and from the public challenge organized by the International Skin Imaging Collaboration (ISIC)~\cite{tschandl2018ham10000}. 
{Path16 is a subset of publicly released histopathology images collated from multiple publicly available datasets ~\cite{Oral_Cancer, patch_camelyon,breast, lc25000,zheng2022deep,mhist}, including eleven diseases and five normal classes (see Supplementary Material for more details about dataset generation). These data are divided into seven tasks based on source of images, including Oral cavity (OR, 2 classes), Lymph node (LY, 2 classes), Breast (BR, 2 classes), Colon (CO, 2 classes), Lung (LU, 2 classes), Stomach (ST, 4 classes), and Colorectal polyp (CP, 2 classes).} 
In training, each image is randomly rotated and then resized to $224\times 224$ pixels.

In all experiments, publicly released CNN models which are pretrained on the Imagenet-1K dataset were used for the fixed feature extractor.
During continual learning, the stochastic gradient descent optimizer was used for task-specific adapter learning, with batch size 32, weight decay 0.0005, and momentum 0.9.  The initial learning rate was $0.01$ and  decayed by a factor of 10 at the 70th, 100th and 130th epoch, respectively. The adapters were trained for up to 200 epochs with consistently observed convergence. For fine-tuning classifier heads, the Adam optimizer was adopted, with initial learning rate 0.001 which decayed by a factor of 10 at the 55th, and 80th, respectively. The classifier heads were fine-tuned for 100 epochs with convergence observed. 
Unless otherwise mentioned, ResNet18 was used as the backbone, the size of memory for storing old images was {40 on Skin8, 80 on Path16, 2000 on CIFAR100 {and 200 on MedMNIST}}. 

\begin{table}[tb]
\centering
\caption{Statistics of three datasets. `[600, 1024]': the range of image width and height. }
\begin{tabular}{c|c|c|c|c|c}
\hline
\textbf{Dataset} & \textbf{Classes} & \textbf{Train set} & \textbf{Test set} & \textbf{Number of tasks} & \textbf{Size} \\ \hline
\textbf{Skin8}\cite{tschandl2018ham10000}           & 8     & 3,555     & 705   & 4     & [600, 1024]  \\
\textbf{Path16}                                     & 16    & 12,808    & 1,607 & 7     & 224 $\times$ 224   \\ 
\textbf{CIFAR100}\cite{krizhevsky2009learning}      & 100   & 50,000    & 10,000   & 5, 10 & 32 $\times$ 32 \\
\textbf{MedMNIST}\cite{medmnistv1}      & 36   & 302,002    & 75,659   & 4 & 28 $\times$ 28 \\
\hline
\end{tabular}
\label{tab1}
\end{table}

In continual learning, the classifier sequentially learned multiple tasks, with each task a small number of new classes (e.g., 2, 10, 20). After learning each task, the mean class recall (MCR) over all classes learned so far is used to measure the classifier's performance. Note that MCR is equivalent to classification accuracy for class-balanced test set. For each experiment, the order of classes is fixed, and all methods were executed three times with different initialization. The mean and standard deviation of MCRs over three runs were reported.


\begin{figure}[t]
\centering
\includegraphics[width=0.3\textwidth,height=0.2\textwidth]{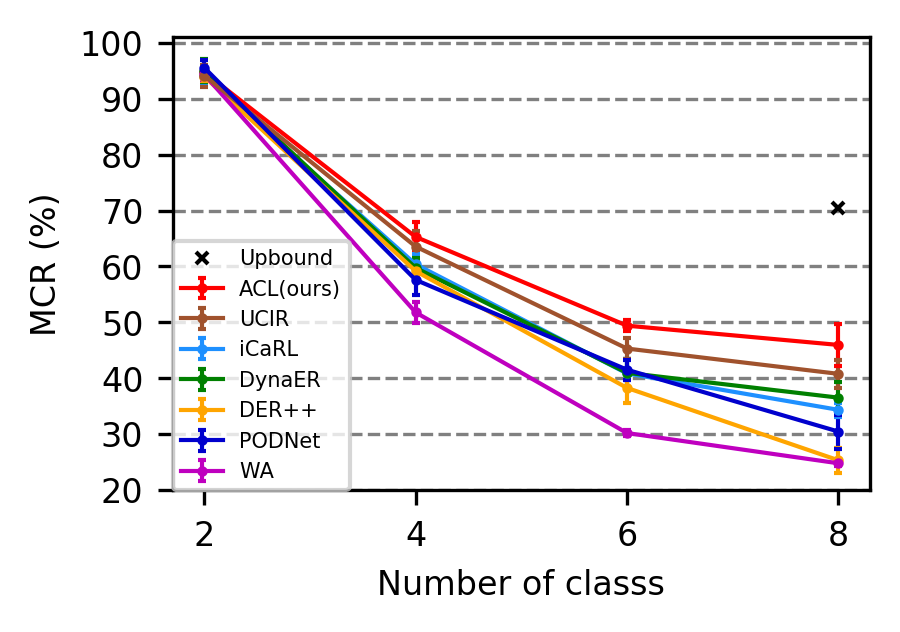}
\includegraphics[width=0.3\textwidth,height=0.2\textwidth]{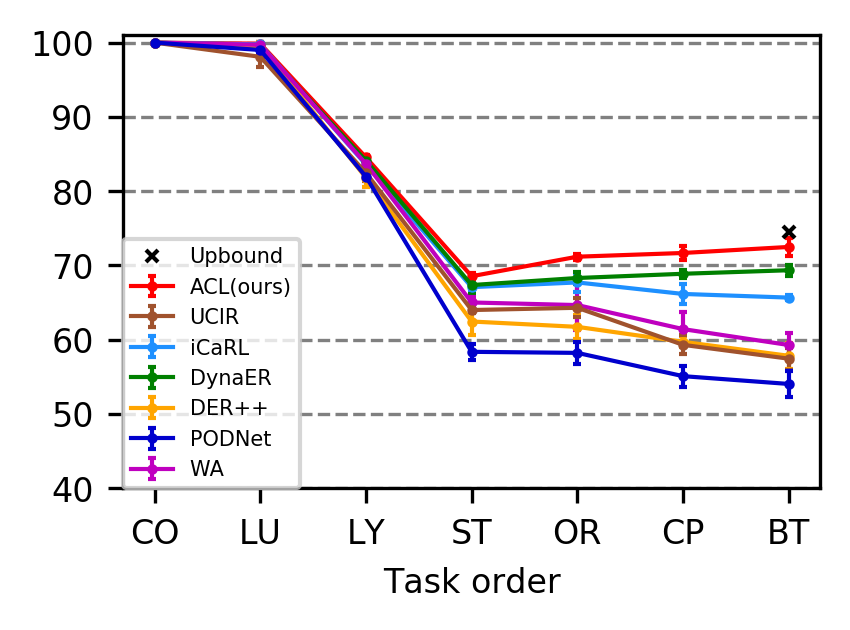}
\includegraphics[width=0.3\textwidth,height=0.2\textwidth]{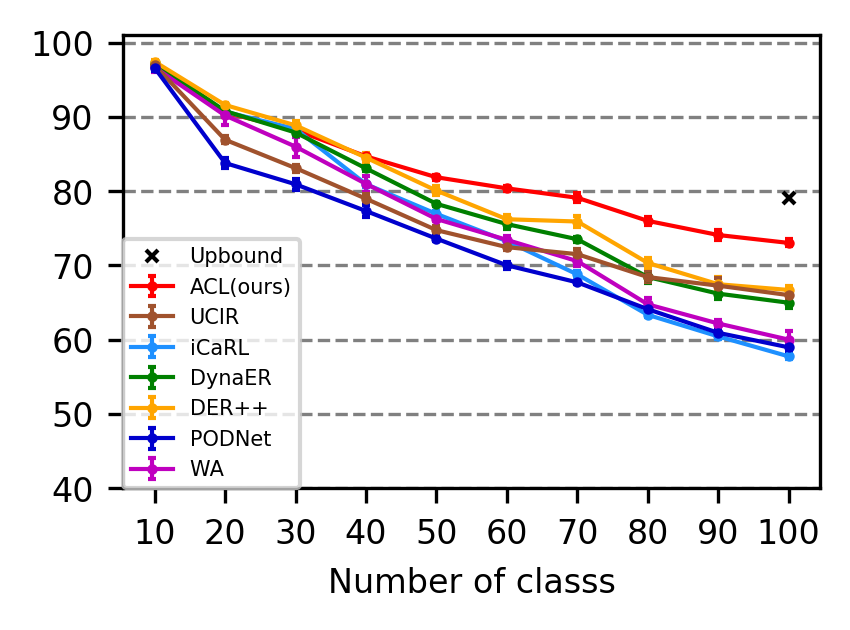}
\includegraphics[width=0.3\textwidth,height=0.2\textwidth]{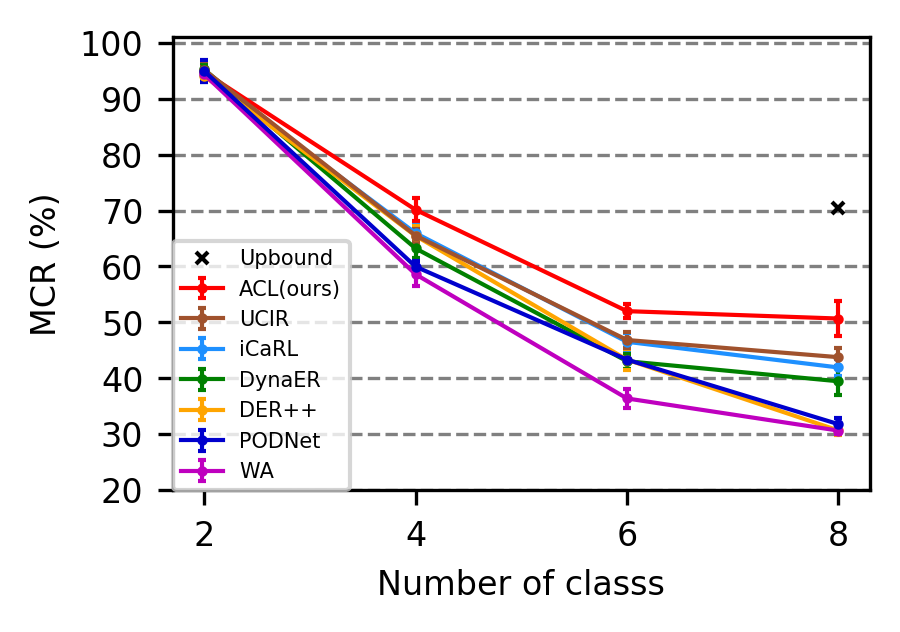}
\includegraphics[width=0.3\textwidth,height=0.2\textwidth]{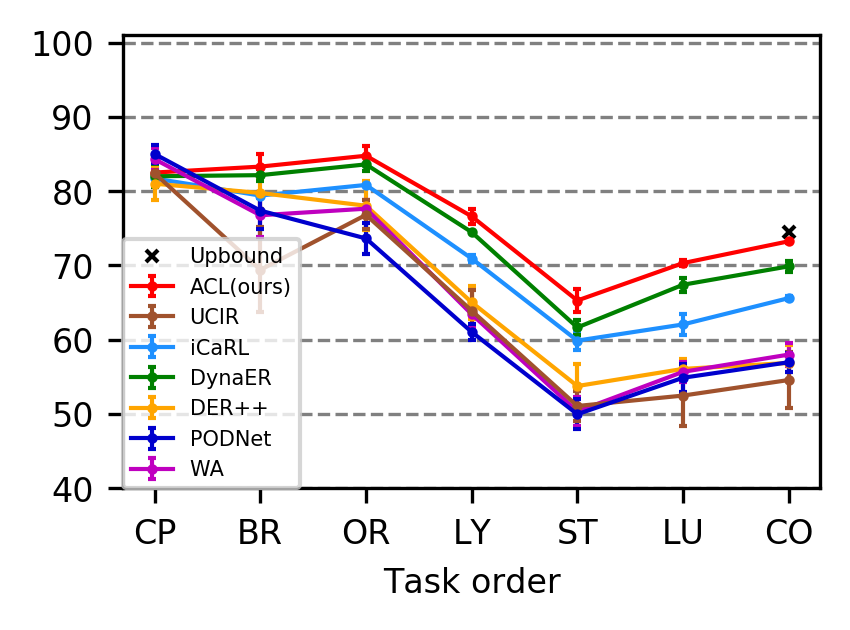}
\includegraphics[width=0.3\textwidth,height=0.2\textwidth]{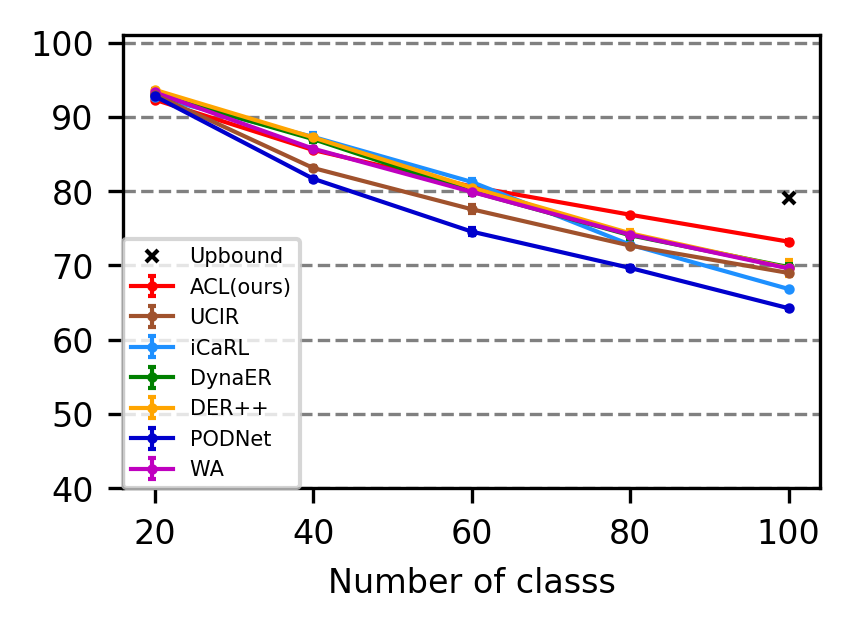}
\caption{
{Performance of continual learning on the Skin8, Path16 and CIFAR100 dataset, respectively. First column: 2 new classes each time on Skin8 respectively with memory size 16 and 40. Second column: continual learning on Path16 in different task orders. Last column: respectively learning 10 and 20 new classes each time on CIFAR100.}
}
\label{fig:effective}
\end{figure}

\subsection{Result analysis}
\noindent\textbf{Effectiveness evaluation}: In this section, we compare {ACL} 
against state-of-the-art baselines, including iCaRL~\cite{rebuffi2017icarl}, DynaER~\cite{yan2021dynamically}, DER++~\cite{buzzega2020dark}, WA~\cite{zhao2020maintaining}, PODNet~\cite{douillard2020podnet}, and UCIR~\cite{hou2018lifelong}. 
In addition, an upper-bound result (from a classifier which was trained with all classes of training data) is also reported. {Similar amount of effort was taken in tuning each baseline method.} As shown in Figure~\ref{fig:effective}, our method outperforms all strong baselines in almost all settings, no matter whether the classifier learn continually 2 classes each time on Skin8 (Figure~\ref{fig:effective}, first column),
in two different task orders on Path16 (Figure~\ref{fig:effective}, second column), in {10 or 20 classes} each time on CIFAR100 (Figure~\ref{fig:effective}, last column){, or in 4 different domains on MedMNIST\cite{medmnistv1} (Figure 1 in Supplementary Material)}.
Note that performance of most methods does not decrease (or even increase) at the last two or three learning rounds on Path16, probably because most methods perform much better on these tasks than on previous rounds of tasks. 


\noindent\textbf{Ablation study}: An ablation study was performed to evaluate the performance gain of each proposed component in {ACL} 
. Table~\ref{tab:ablation} (first four rows) shows that the continual learning performance is gradually improved while more components are included, confirming the effectiveness of each proposed component. In addition, when fusing all the task features with a unified classifier head (Table~\ref{tab:ablation}, last row), the continual learning performance is clearly decreased compared to that from the proposed method (fourth row), 
confirming the effectiveness of task-specific classifier heads for class-incremental learning.


\begin{table}[t]
\centering
\caption{Ablation study of {ACL} 
on Skin8 (with 2 new class per time) and on CIFAR100 (with 10 new classes per time). `T.S.H.': inclusion of task-specific heads; `Others': inclusion of the `others' output neuron in each head. `Avg': average of MCRs over all rounds of continual learning; `Last': MCR at the last round.
}
\resizebox{\textwidth}{!}{
\setlength{\tabcolsep}{1mm}{
\begin{tabular}{cccccccc}
\hline
\multicolumn{4}{c}{\textbf{Components}} & \multicolumn{2}{c}{\textbf{Skin8}} & \multicolumn{2}{c}{\textbf{CIFAR100}} \\
\multicolumn{1}{c}{\textbf{T.S.H}} &
  \multicolumn{1}{c}{\textbf{Adapter}} &
  \multicolumn{1}{c}{\textbf{Others}} &
  \textbf{Fine-tune} &
  \multicolumn{1}{c}{\textbf{Avg}} &
  \multicolumn{1}{c}{\textbf{Last}} &
  \textbf{Avg} &
  \multicolumn{1}{c}{\textbf{Last}} \\ \hline
\Checkmark    &     &   &   & $50.91_{\pm 0.18}$   & $27.47_{\pm 0.32}$   &  41.68$_{\pm 0.04}$ & 18.64$_{\pm 0.14}$\\ 
\Checkmark    & \Checkmark      &   &   & $60.89_{\pm 0.56}$   & $35.4_{\pm 1.20}$   &  47.22$_{\pm 0.09}$ & 21.20$_{\pm 0.13}$\\
\Checkmark    & \Checkmark      & \Checkmark    &   & $60.90_{\pm 1.97}$   & $42.18_{\pm 2.65}$   & 58.72$_{\pm 0.07}$  & 46.44$_{\pm 0.42}$\\
\Checkmark    & \Checkmark      & \Checkmark    & \Checkmark    & \textbf{66.44}$_{\pm 0.90}$   & \textbf{50.38}$_{\pm 0.31}$   &  \textbf{82.50}$_{\pm 0.39}$ & \textbf{73.02}$_{\pm 0.47}$\\ 
    & \Checkmark      &     & \Checkmark  & $64.80_{\pm 0.87}$   & $46.77_{\pm 1.58}$   &  81.67$_{\pm 0.39}$ & 70.72$_{\pm 0.33}$\\ \hline
\end{tabular}}}
\label{tab:ablation}
\end{table}

\begin{table}[t]
\centering
\caption{Continual learning performance with different CNN backbones. 
Two new classes and ten new classes were learned each time on Skin 8 and CIFAR100, respectively. The range of standard deviation is [0.06, 3.57].}
\resizebox{\textwidth}{!}{
\begin{tabular}{cl|ccc|ccc|ccc}
\hline
\multicolumn{2}{c|}{\textbf{Backbones}} &
  \multicolumn{3}{c|}{\textbf{ResNet18}} &
  \multicolumn{3}{c|}{\textbf{EfficientNet-B0}} &
  \multicolumn{3}{c}{\textbf{MobileNetV2}} \\ \hline
\multicolumn{2}{c|}{\textbf{Methods}} &
  \textbf{iCaRL} &
  \textbf{DynaER} &
  \multicolumn{1}{l|}{\textbf{{ACL(ours)}}}  &
  \textbf{iCaRL} &
  \textbf{DynaER} &
  \multicolumn{1}{l|}{\textbf{{ACL(ours)}}}  &
  \textbf{iCaRL} &
  \textbf{DynaER} &
  \multicolumn{1}{l}{\textbf{{ACL(ours)}}} \\ \hline
\multirow{2}{*}{\textbf{Skin8}} &
  \multicolumn{1}{c|}{\textbf{Avg}} &
  62.16 &  60.24 &  \textbf{66.44} &     
  61.17 &  60.52 &  \textbf{66.86} &     
  64.58 &  62.16 &  \textbf{66.08} \\     
 &
  \textbf{Last} &
  41.94 &  39.47 &  \textbf{50.38} &      
  42.60 &  40.17 &  \textbf{48.50} &      
  42.52 &  41.49 &  \textbf{48.83} \\ \hline      
\multirow{2}{*}{\textbf{CIFAR100}} &
  \multicolumn{1}{c|}{\textbf{Avg}} &
  75.74 &  78.59 &  \textbf{82.50} &     
  73.98 &  81.98 &  \textbf{84.56} &     
  73.47 &  77.04 &  \textbf{81.04} \\    
 &
  \textbf{Last} &
  57.75 &  64.97 &  \textbf{73.02} &     
  53.40 &  70.13 &  \textbf{75.55} &     
  55.00 &  64.30 &  \textbf{70.88} \\ \hline 
\end{tabular}}
\label{tab:generalization}
\end{table}

\noindent\textbf{Generalizability study}: The pretrained feature extractor with different CNN backbones were used to evaluate the generalization of {ACL} 
. As shown in 
Table~\ref{tab:generalization}, 
{ACL} consistently outperforms representative strong baselines with each CNN backbone {(ResNet18\cite{he2016deep}, EfficientNet-B0\cite{tan2019efficientnet} and MobileNetV2\cite{sandler2018mobilenetv2}) on Skin8},  supporting the generalizability of our method.

\section{Conclusion}
Here we propose a new adapter-based strategy for class-incremental learning of new diseases. The learnable light-weight and task-specific adapters, together with the pretrained and fixed feature extractor, can effectively learn new knowledge of diseases and meanwhile keep old knowledge from catastrophic forgetting. The task-specific heads with the special `out-of-distribution' output neuron within each head  helps keep extracted features discriminative between different tasks. Empirical evaluations on multiple medical image datasets confirm the efficacy of the proposed method. We expect such adapter-based strategy can be extended to other continual learning tasks including lesion detection and segmentation.

\noindent\textbf{Data Use Declaration}: Our experimental data were collected from open source datasets. 
{The Skin8, MedMNIST, CIFAR100 and Path16 can be downloaded at: \href{https://challenge.isic-archive.com/data/\#2019}{https://challenge.isic-archive.com/data/\#2019}, \href{https://medmnist.com/}{https://medmnist.com/}, \href{https://www.cs.toronto.edu/~kriz/cifar.html}{https:\newline //www.cs.toronto.edu/~kriz/cifar.html} and \href{https://drive.google.com/drive/folders/1LMHgawD83Z5EmYN6wtLVIibZNrAZglZt?usp=sharing
}{https://drive.google.com/drive/fold\newline 
ers/1LMHgawD83Z5EmYN6wtLVIibZNrAZglZt?usp=sharing}, respectively.}

\noindent{\textbf{Acknowledgement.} This work is supported in part by
the Major Key Project of PCL (grant No. PCL2023AS7-1), the National Natural Science Foundation of China (grant No. 62071502 {\& No. 12071496), Guangdong Excellent Youth Team Program (grant No. 2023B1515040025), and the Guangdong Provincial Natural Science Fund (grant No. 2023A1515012097)}.
}
\bibliographystyle{splncs04}
\bibliography{ref}
%





\end{document}


\title{Adapter Learning in Pretrained Feature Extractor for Continual Learning of Diseases [Supplementary Material]}
%
%
%
\author{Wentao Zhang\inst{1,4}\thanks{Authors contributed equally.} \and 
Yujun Huang\inst{1,4}$^\star$ \and 
Tong Zhang\inst{2} \and 
Qingsong Zou\inst{1,3} \and 
Wei-Shi Zheng\inst{1,4} \and 
Ruixuan Wang\inst{1,2,4}\thanks{Corresponding author: wangruix5@mail.sysu.edu.cn}} 

%

\institute{School of Computer Science and Engineering, Sun Yat-sen Univerisity, Guangzhou, China  \and Peng Cheng Laboratory, Shenzhen, China \and Guangdong Province Key Laboratory of Computational Science, Sun Yat-sen University, Guangzhou, China \and Key Laboratory of Machine Intelligence and Advanced Computing, MOE, Guangzhou, China}
\maketitle              

\begin{table}
\centering
\caption{Statistics of Path16. 
{For Oral cancer, PatchCamelyon, LC25000, and MHIST datasets, approximately 800, 400, 800, and 500 training images per class were randomly selected from the original training set, respectively. Additionally, around 100 testing images per class were randomly chosen from each original test set.}
For the Breast cancer and the TCGA-STAD datasets, the original datatasets were first organized at the patient or slice level, and then around 800 training and 100 test patches were randomly cropped from the training and test slices, with similar number of training/test patches cropped from each training/test slice. `[224, 2048]’: the range of image width and height}
\label{tab1}
\setlength{\tabcolsep}{1mm}{
\begin{tabular}{c|c|c|c|c|c}
\hline
Body parts          &  Source dataset                       &  Classes  & Train set     & Test set   & Size         \\
\hline
Oral cavity (OR)    & Oral cancer~[12]        & 2         & 1,641         &  200         & $[224, 2048]$  \\
Lymph node (LY)     &PatchCamelyon~[24]    & 2         & 800         &  200         & $96 \times 96$ \\
Breast (BR)         & Breast cancer~[6]          & 2         & 1,600         &  200         & $50 \times 50$\\
Colon (CO)          & LC25000~[1]                & 2         & 1,600         &  200         & $768 \times 768$  \\
Lung (LU)           & LC25000~[1]               & 2         & 1,600         &  200         & $768 \times 768$ \\
Stomach (ST)        & TCGA-STAD~[32]                & 4         & 3,208         & 407          & $224 \times 224$\\
Colorectal polyp (CP)     & MHIST~[28]             & 2         & 1,000         &  200         & $224 \times 224$\\

\hline
\end{tabular}}
\end{table}


\begin{figure}[h]
\centering
\includegraphics[width=0.6\textwidth,height=0.4\textwidth]{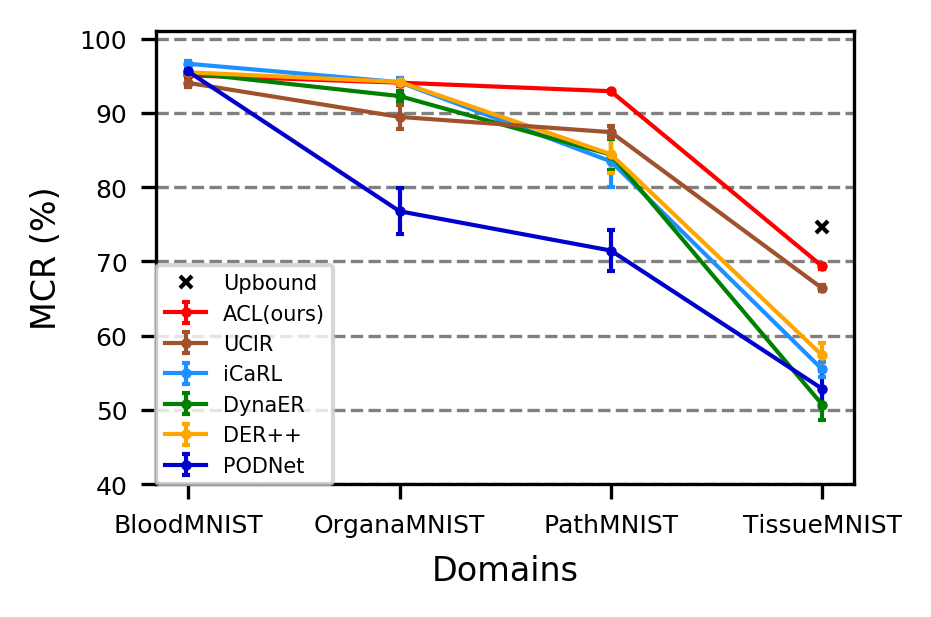}

\caption{
{Performance of continual learning over 4 different domains on MedMNIST dataset. 
}
}
\label{fig:effective}
\end{figure}

\bibliographystyle{splncs04}
%



